\documentclass[letterpaper, 10 pt, conference]{ieeeconf}  

\IEEEoverridecommandlockouts                              

\usepackage{multirow}
\usepackage{amsmath}
\usepackage{booktabs}
\usepackage{graphicx}
\usepackage{hyperref}

\hypersetup{hypertex=true,
            colorlinks=true,
            linkcolor=black,
            anchorcolor=black,
            citecolor=black}

\usepackage{bm}
\usepackage{algorithmicx,algorithm}
\usepackage[noend]{algpseudocode}

\title{\LARGE \bf
Mask-Guided Image Person Removal with Data Synthesis
}

\author{Yunliang Jiang$^{1}$, Chenyang Gu$^{1,2}$, Zhenfeng Xue$^{2,3,*}$, Xiongtao Zhang$^{1}$, Yong Liu$^{2,3,*}$
\thanks{$^{1}$Yunliang Jiang, Chenyang Gu and Xiongtao Zhang are with School of Information Engineering, Huzhou University, Huzhou, China.}%
\thanks{$^{2}$Chenyang Gu, Zhenfeng Xue and Yong Liu are with Intelligent Perception and Control Center, Huzhou Institute of Zhejiang University, Huzhou, China. Corresponding author: Zhenfeng Xue (zfxue0903@zju.edu.cn), Yong Liu (yongliu@iipc.zju.edu.cn)}%
\thanks{$^{3}$Zhenfeng Xue and Yong Liu are with Institute of Cyber-Systems and Control, Zhejiang University, Hangzhou, China}%
}

\begin{document}

\maketitle
\thispagestyle{empty}
\pagestyle{empty}

\begin{abstract}

As a special case of common object removal, image person removal is playing an increasingly important role in social media and criminal investigation domains.
Due to the integrity of person area and the complexity of human posture, person removal has its own dilemmas.
In this paper, we propose a novel idea to tackle these problems from the perspective of data synthesis.
Concerning the lack of dedicated dataset for image person removal, two dataset production methods are proposed to automatically generate images, masks and ground truths respectively.
Then, a learning framework similar to local image degradation is proposed so that the masks can be used to guide the feature extraction process and more texture information can be gathered for final prediction.
A coarse-to-fine training strategy is further applied to refine the details.
The data synthesis and learning framework combine well with each other.
Experimental results verify the effectiveness of our method quantitatively and qualitatively, and the trained network proves to have good generalization ability either on real or synthetic images.

\end{abstract}

\maketitle

\section{Introduction}~\label{Sec:one}

These years have witnessed the development of machine learning and image processing, and it also brings difficulties to incident or criminal investigation, especially when the images are easy to be faked.
With the development of smart phones and cameras, there is a strong trend to handle digital images on social media, and sometimes they often do not wish some specific persons to appear in the images.
On the contrary, the image deep fake technology of person has important reference significance for criminal investigation in the field of public security.
Hence, person removal~\cite{bharathi2021real,yae2022people} in digital images is opening up an emerging and interesting research domain.

As a branch of common object removal~\cite{lim2021erasor,bevsic2022dynamic}, image person removal has its own difficulties.
Different from object removal, person removal focuses on the removal of specific categories with complete area and the posture of person is richer than that of the common objects. 
This makes traditional object removal methods perform bad on removing persons, and the datasets need be to expanded purposefully.
Compared with watermark removal~\cite{chen2021refit,liu2021wdnet}, the person has a three-dimensional posture, while the watermark is only two-dimensional.
This may lead to the loss of some key spatial information in the process of feature extraction if we simply apply the methods of watermark removal.

\begin{figure}[t]
	\centering
	\includegraphics[scale=0.42]{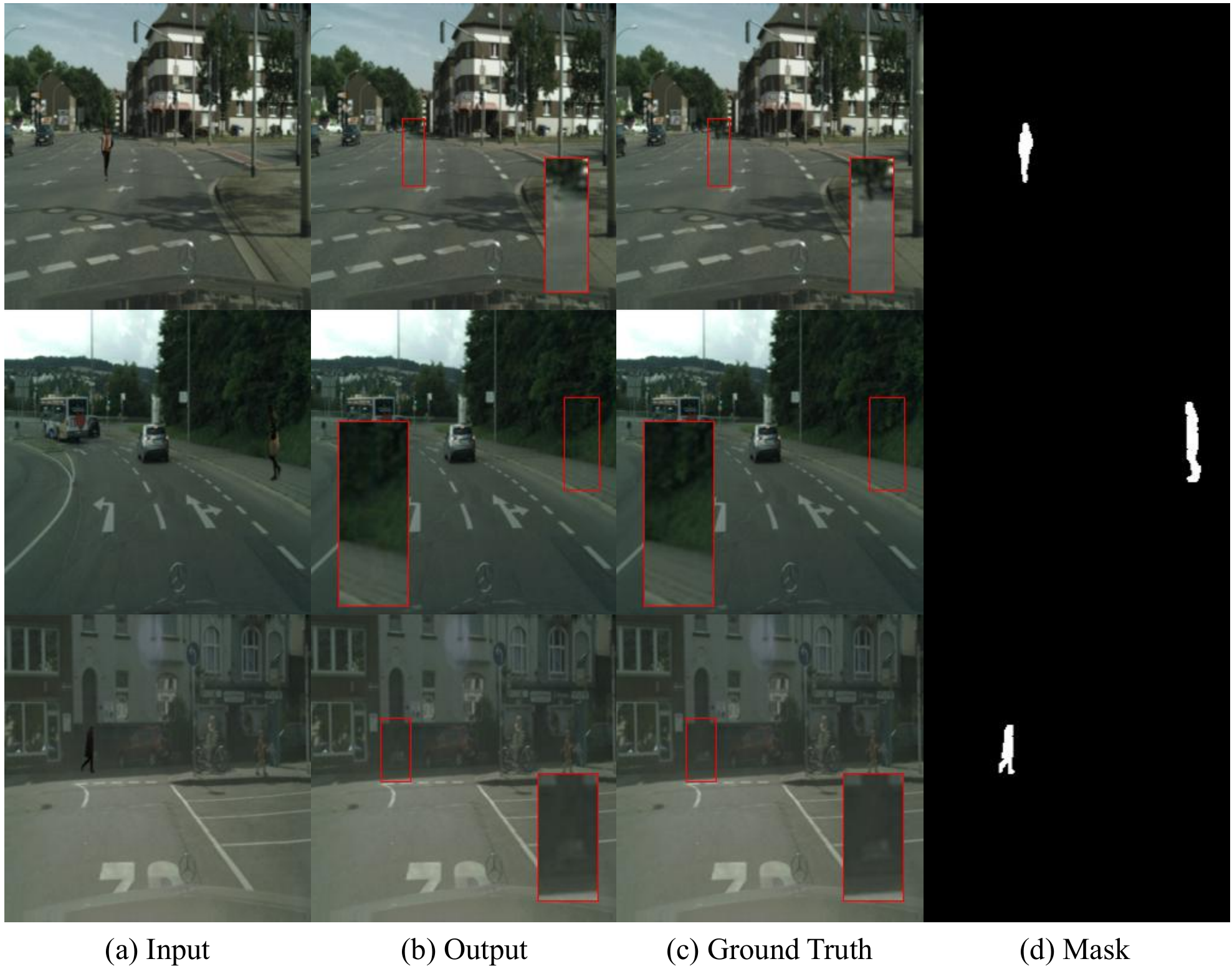}
	\caption{Visualization of image person removal.}
	\label{fig:overview}
\end{figure}

Currently, due to the mature development of common object removal methods, we mainly focus on improving person removal from the perspective of dataset simulation and learning framework.
On the one hand, the traditional methods of making dataset by hand is usually time-consuming and laborious, and moreover it is hard to obtain two images with and without person area simultaneously.
Therefore, we propose two methods of dataset production to improve the efficiency and accuracy of person removal task  with the help of data synthesis technology~\cite{sun2019dissecting,yao2020simulating,xue2021learning,xue2022image}.
This enables us to obtain images with person area, ground truths without person area and masks of person area simultaneously.
On the other hand, we make improvement on the learning framework of image person removal.
Without image and ground truth pair, traditional methods~\cite{shetty2018adversarial,yae2022people,liu2021wdnet,gao2022towards} focus on image inpainting after the person area is subtracted from original image.
In our method, the image pair is accessible and we propose a novel learning framework similar to local image degradation with an $\alpha$ blending effect.
This enables the local person areas fade away gradually, and the background is filled.
The person mask is used to guide the feature extraction process, and a coarse-to-fine learning strategy is applied to refine the details.
The data synthesis and learning framework combine well with each other and it achieves good performance.
Some visual examples are presented in Fig.~\ref{fig:overview}.

Specifically, one of the data synthesis method is called the image mosaic method. In this method, two real-world datasets are selected as source materials, including one with person masks such as Daimler Pedestrian Detection Benchmark~\cite{flohr2013pedcut}, and another with street backgrounds such as Cityscapes~\cite{cordts2016cityscapes} and BDD100K~\cite{yu2020bdd100k}.
Next, we randomly select an image from each of the two datasets, and subtract the person area of the first image using the mask, and place it on the second background image to form a new street view image with a person mask.
The position of person is manually adjusted by us to make it roughly conform to the structure of the scene.
This enables us to quickly obtain a dataset dedicated to person removal.
The original, target and mask images are obtained simultaneously.

The second is the virtual rendering method. In this method, one real-world dataset is selected as the background images, and we collect some virtual person models in rendering engine from existing projects~\cite{sun2019dissecting}.
With the help of rendering engine like Unity or Unreal, we are able to put the three-dimensional model of the virtual person into the real background image to generate a realistic image.
The benefits of this approach are manifold. Firstly, the position and posture of the person model in the image can be manually controlled and changed arbitrarily.
Thus at the same position, person images with different poses and angles can be generated, which is beneficial for neural networks to extract abundant features.
Secondly, the lighting of the person model can be arbitrarily edited, which can make the generated person lighting model more realistic, while the above method can not change the lighting.
Thirdly, a variety of person masks can be automatically generated by programming, including detection, segmentation as well as depth.
This allows us to quickly obtain datasets with rich content and illumination. Finally, we design a learning framework to adjust the lighting condition.

We've performed extensive experiments to verify the effectiveness of the method.
The contributions are as follows.

\begin{itemize}
	\item Two data synthesis methods are designed specifically for the image person removal task, so that original, target and mask images are obtained simultaneously.
	
	\item A novel mask-guided learning framework similar to local image degradation with a coarse-to-fine strategy is proposed to boost the removal accuracy.
	
	\item Extensive experiments are performed on our dataset and method, and the two data synthesis methods show good generalization ability on third-party images.

\end{itemize}

The rest of this paper is arranged as follows. In Sect.~\ref{sec:2}, we review some related works.
The dataset synthesis methods are introduced in Sect.~\ref{sec:3}. In Sect.~\ref{sec:4}, we introduce the learning framework, and experimental results are analysed in Sect.~\ref{sec:5}. Finally, we make a summary in Sect. 6.

\section{Related works}~\label{sec:2}

\subsection{Common object removal} 

The person removal task has not been widely studied yet, and it can be viewed as a special case of common object removal.
Existing object removal methods mainly focusing on recovering the images after the specific areas to be removed are subtracted from the original images, \textit{i.e.}, the so-called image inpainting task.
At present, there are two mainstream methods for image inpainting, including the CNN-based and the GAN-based methods.
The CNN-based methods use to design advanced networks to predict the damaged area.
Jain \textit{et al.}~\cite{jain2008natural} becomes one of the pioneers to use CNN for image inpainting. The authors use the parameter learning back propagation formula to remove the noise in the image.
Xie \textit{et al.}~\cite{xie2012image} improve the method and propose a denoising autoencoder to handle inpainting of inconsistent locations of damaged pixels.
Cai \textit{et al.}~\cite{cai2017blind} propose a blind image inpainting method using encoder-decoder network structure.
Sidorov \textit{et al.}~\cite{sidorov2019deep} further propose a network architecture for denosing, inpainting and super-resolution for noised, inpainted and low-resolution images respectively.
The GAN-based methods use to recover the images with an adversarial training stages so that the generated images look similar to that of original images.
Pathak \textit{et al.}~\cite{pathak2016context} pioneer the combination of contextual encoder structure with generative adversarial networks~\cite{goodfellow2020generative} for image inpainting, named CE.
Lizuka \textit{et al.}~\cite{iizuka2017globally} further propose a contextual local discriminator and introduce a dilated convolutional layer to increase the receptive field of CE.
Against the shortcoming of simple encoder structure, Liu \textit{et al.}~\cite{liu2020rethinking} propose the MEDFE that use multi-scale and texture information to balance image structure and texture feature consistency.
Yan \textit{et al.}~\cite{yan2018shift} propose the Shift-Net to replace the fully connected layer with a shift connection layer based on U-Net~\cite{ronneberger2015u} structure to inpaint images.
Guo \textit{et al.}~\cite{guo2021image} propose the CTSDG that divides the image inpainting task into two interacting subtasks including structure-constrained texture synthesis and texture-guided structure reconstruction.
These works all face to common object removal, and of course enables to removal persons in the images.

\begin{figure*}[t]
	\centering
	\includegraphics[scale=0.52]{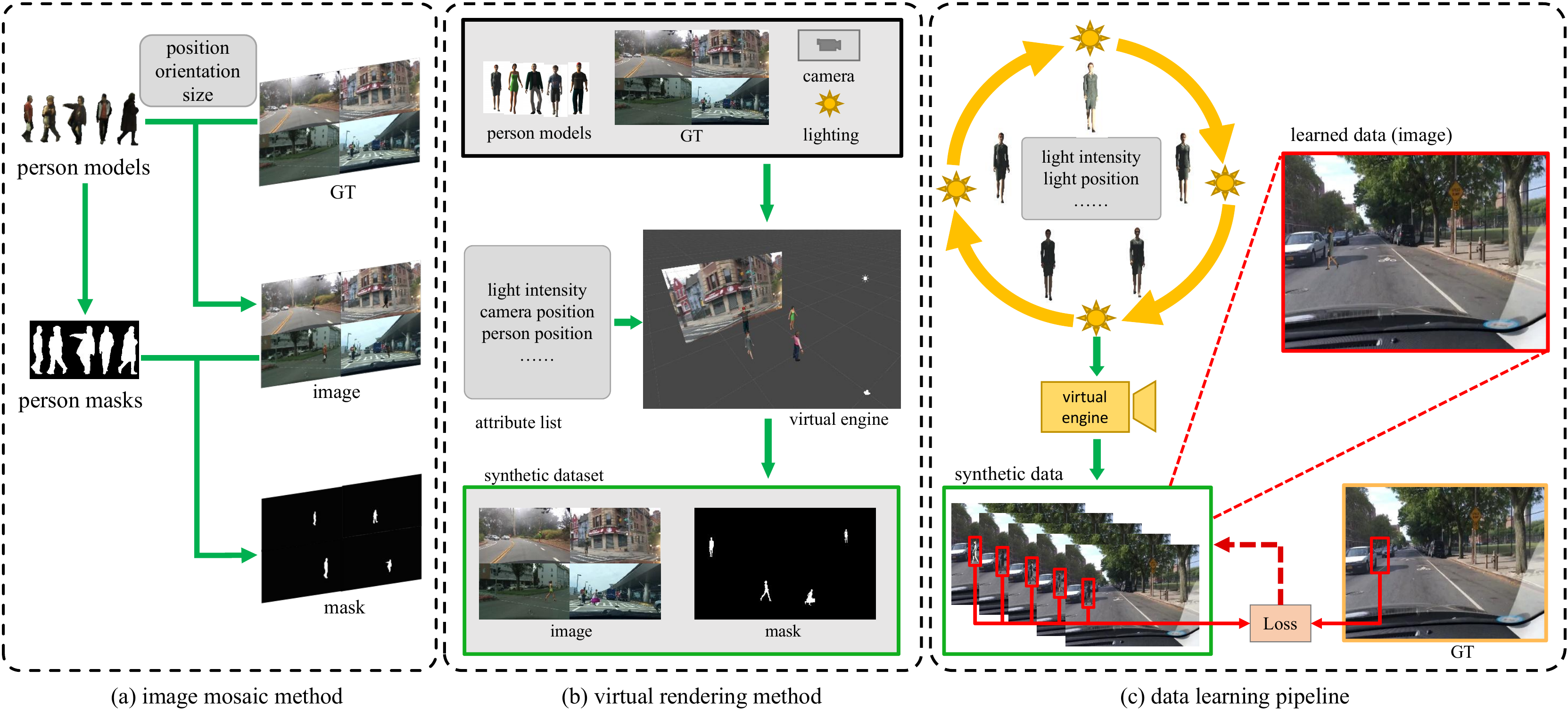}
	\caption{An illustration of the kind of data synthesis method for person removal.}
	\label{fig:data_synthesis}
\end{figure*}

\subsection{Dataset synthesis} 

Due to the complexity of the dataset production process, lastest works begin to concentrate on making dataset through data synthesis, because of the rapidity of label generation in this method.
There are also two mainstream kinds of data synthesis methods, including the image mosaic and the virtual rendering methods.
The image mosaic methods make a new dataset by combining different existing datasets.
Georgakis \textit{et al.}~\cite{georgakis2017synthesizing} synthesize the data by determining the placeable object area and the object zoom value, and then stitching random objects and poses into the real background image.
Dwibedi \textit{et al.}~\cite{dwibedi2017cut} randomly stitch images and perform different mixing steps on the stitched objects, thereby reducing the impact of direct stitching artifacts on training.
Tripathi \textit{et al.}~\cite{tripathi2019learning} propose that the stitched images are generated by a synthesis network and the discriminator is used to synthesize more realistic data.
The virtual rendering methods usually use a rendering engine such as Unity or Unreal to construct a virtual environment, and then render a new dataset with the help of the engine.
Sun \textit{et al.}~\cite{sun2019dissecting} propose the PersonX dataset for person re-identification that contains 1,266 3D characters rendered in Unity.
Yao \textit{et al.}~\cite{yao2020simulating} further extend the engine to vehicle re-identification and propose the attribute descent to optimize the attribute.
Xue \textit{et al.}~\cite{xue2021learning} combine the assets and extend the engine for scene semantic segmentation and propose the SDR method to optimize the scenes.
Hou \textit{et al.}~\cite{hou2022enhancing} propose the CrowdX for enhancing the level of crowd counting algorithms.
For a specific task, Xue \textit{et al.}~\cite{xue2022virfd} apply the data synthesis method to rock size counting task in an engineering problem~\cite{xue2021rock}, leading to improved real-world performance.
While these methods all concentrate on dataset-level data synthesis, Xue \textit{et al.}~\cite{xue2022image} further propose the image-level method to tackle the problem that the attributes are hard to be optimized and use gradient descent to complete the process.
These works show the great potential of data synthesis in various tasks.

\section{Data synthesis method}~\label{sec:3}

Training samples are of vital significance to computer vision tasks.
So far, there is no specific dataset for person removal.
Most of the existing methods are trained on the common object removal benchmarks, and they are generalized to person removal as well. 
However, different from common object removal that use to recover a subtracted image to a complete one, person removal aims to removal specific category in pre-defined area.
It usually wishes to recover the whole person area, and moreover the persons have rich textures and postures.
And thus, there exists a strong domain gap between common object removal and person removal tasks.
Directly applying neural networks trained on common datasets behave not so good on dedicated person removal datasets.

One straightforward way is to collect and annotate person removal dataset, which consumes time and energy.
In this paper, we propose to make person removal datasets from the perspective of data synthesis.
Two methods are introduced here, with one called the image mosaic method, and another called the virtual rendering method.

\subsection{Image mosaic}

In this data synthesis method, two datasets are required, including one to prepare person images and masks~\cite{flohr2013pedcut}, and one to prepare backgrounds~\cite{cordts2016cityscapes,yu2020bdd100k}.
An illustration of the image mosaic process is shown in Fig.~\ref{fig:data_synthesis} (a).
A background image is firstly selected.
Then, a source image is selected and the person model is cropped out using the person mask.
Finally, the person model is attached to the background image, resulting in the target image, as well as the person mask.

In order to make the mosaic image look more realistic, we manually modify the position of the person model in the target image, so that the person is basically in a reasonable position, such as on the road or on the crosswalk.
A total number of 50 background images and 10 person model images are selected, and they are randomly combined, resulting in 500 images.
We can this dataset the real person removal dataset (Real Synthesis PR Benchmark), because the dataset is made all by real-world images.

\subsection{Virtual rendering}

There is an obvious problem in the dataset generated by the above method, that is, the person model is monotonous. 
The reason is that the person model extracted from the image has fixed lighting and human posture. 
This makes it difficult to change the texture of the person model and the movements of the feet, which may lead to the authenticity of the content diversity of the generated dataset.

Therefore, we propose the second method, i.e., the virtual rendering method. 
In this method, one real dataset is used to provide background images, and a virtual rendering engine such as Unity is used to generate person models. 
The material library of the virtual person model can be found in the open source project~\cite{sun2019dissecting}.

The principle of this method is shown in Fig.~\ref{fig:data_synthesis} (b). 
Firstly, one background image is selected and imported into the rendering engine, and we place it virtually in front of the scene camera. 
Then, one virtual person model is placed in front of the background image. Within the engine, the attributes of virtual person model is editable, including the position, posture, scale and etc. 
As before, we artificially set some positions for the characters in each background image, so that the person can appear in the correct position in the image, such as sidewalk and zebra crossing. 
Finally, the attributes for lighting are modified, so that the texture of person model can be adjusted to fit the environment.

There are several advantages for this method.
With the virtual engine, the person model and lighting conditions can be arbitrarily modified to increase the degree of coincidence between the person model and the background image. 
Meanwhile, the posture of the person model in the generated dataset is more abundant, and the content of the dataset is also more abundant. 
Last but not least, the person model annotation, not only the mask, can be automatically generated by the rendering buffer of the engine, so the content of the annotation is more abundant.
Thus, the dataset made in this way contains more modal and texture information.

Like~\cite{sun2019dissecting,yao2020simulating,xue2021learning}, we connect the Python programming and virtual engine through ML-Agents. 
In the Python programming, we set several attribute values of lighting in advance. 
By this way, it enables to adjust the lighting of person model through Python programs. 
The attributes can be randomly initialized, violent enumerated or optimized through learning. 
Here, as shown in Fig.~\ref{fig:data_synthesis} (c), we propose an objective function for illumination learning through the following equation
\begin{equation}
	Loss = L_1 (I_{m}, GT_{m})
\end{equation}
where $I_{m}$ denotes masked the person model area mask, and $GT_{m}$ denotes the masked ground truth, and the $L_1$ loss function is applied.
This objective function forces the texture of the generated person model to be as similar as possible to the texture of the background image.

\section{Person removal pipeline}~\label{sec:4}

\begin{figure*}[t]
	\centering
	\includegraphics[scale=0.8]{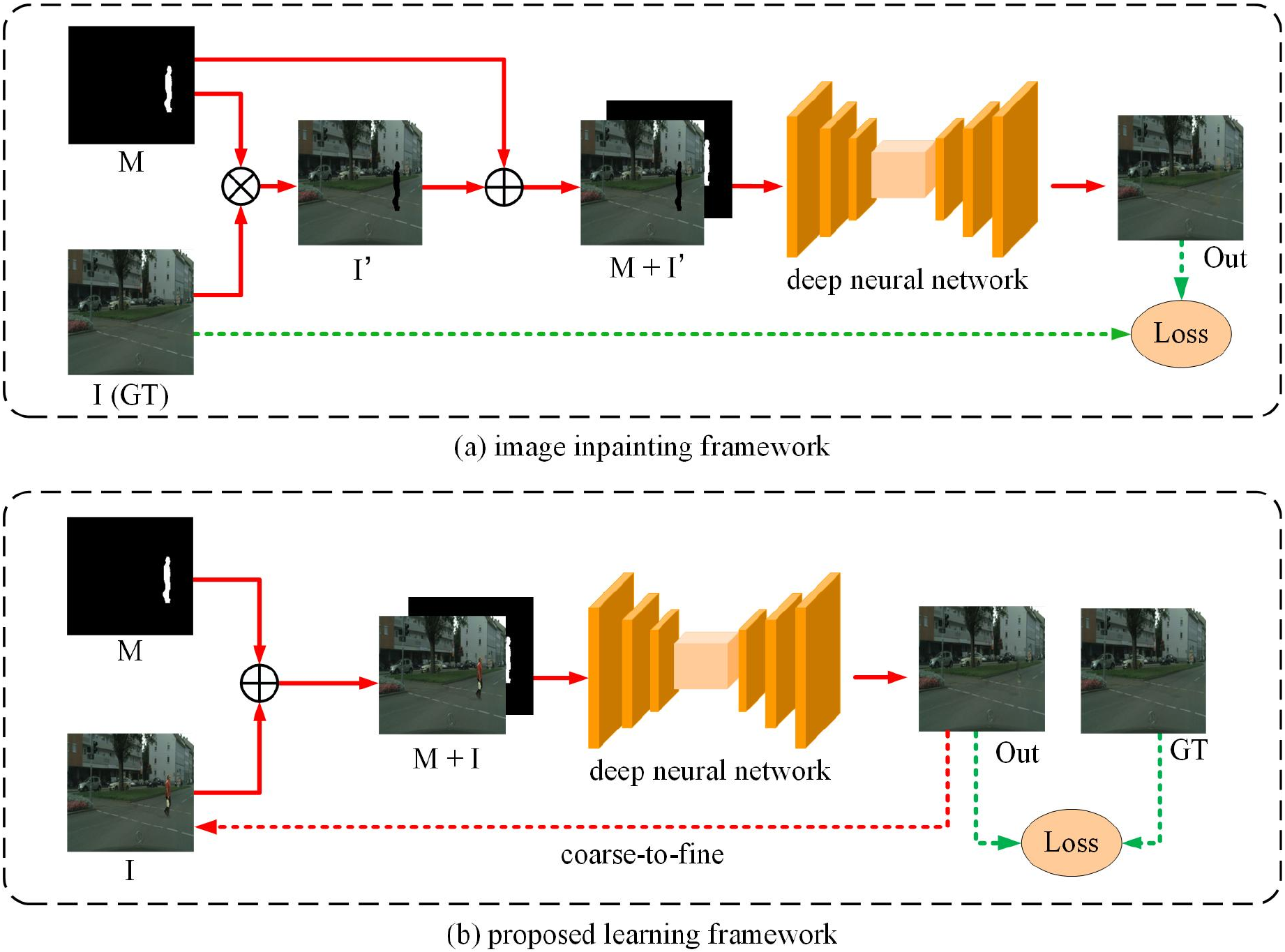}
	\caption{Comparison between image inpainting framework for common object removal and the proposed learning framework for image person removal.}
	\label{fig:method_compare}
\end{figure*}

\subsection{Problem definition}

The objective of image person removal is to restore a clean target image $T$ from a messy source image $I$ with a pixel-level person model binary mask $m$, where $T$, $I$ and $m$ have the same resolution.
Each pixel in the person mask $m$ takes value from `$0$' and `$1$', with `$1$' indicating the person area to be removed.
Given a person removal network $G$, the source image $I$ and person mask $m$ are sent into the network to generate the predictive target image $\hat{T}$.
\begin{equation}
	\hat{T} = f_G(I,m),
\end{equation}
where the corresponding person model in source image $I$ will be removed and replaced with similar background texture around the surrounding environment.
The network parameters are updated with the following loss function
\begin{equation}
	Loss = L_p(\hat{T}, T) + L_1(\hat{T}, T),
\end{equation}
where $L_p$ and $L_1$ denote the perception and regularization loss correspondingly.

Traditionally, as shown in Fig.~\ref{fig:method_compare} (a), object removal follows an image inpainting framework.
It applies the person mask $m$ to the source image $I$, and directly subtracts the person area, resulting in a intermediate image $I'$.
\begin{equation}
	I' = I \otimes m,
\end{equation}
where $\otimes$ denotes the subtracting operation. Note that due to the subtraction process, the input image is arbitrary either with or without person area, so we directly put the ground truth image as input to be conspicuous.
Then $I'$ and $m$ is sent into an image inpainting network $G$, and the final prediction can be represented by the following equation
\begin{equation}
	\hat{T} = (1-m) \otimes I + m \otimes G(I',m).
\end{equation}
Such a learning pipeline divide the person removal task into two independent stages, including person subtraction and image restoration.
This approach has many problems.
First, the person subtraction stage will completely erase the color information of the person area, which leads to the loss of dimensions during image restoration process.
Second, the matching degree with the surrounding texture is not well considered in the process of image restoration.
Therefore, the image generated in this way has poor quality, and there always exists serious blur, ghost and boundary.

\subsection{Proposed learning pipeline}

\begin{figure}[t]
	\centering
	\includegraphics[scale=0.35]{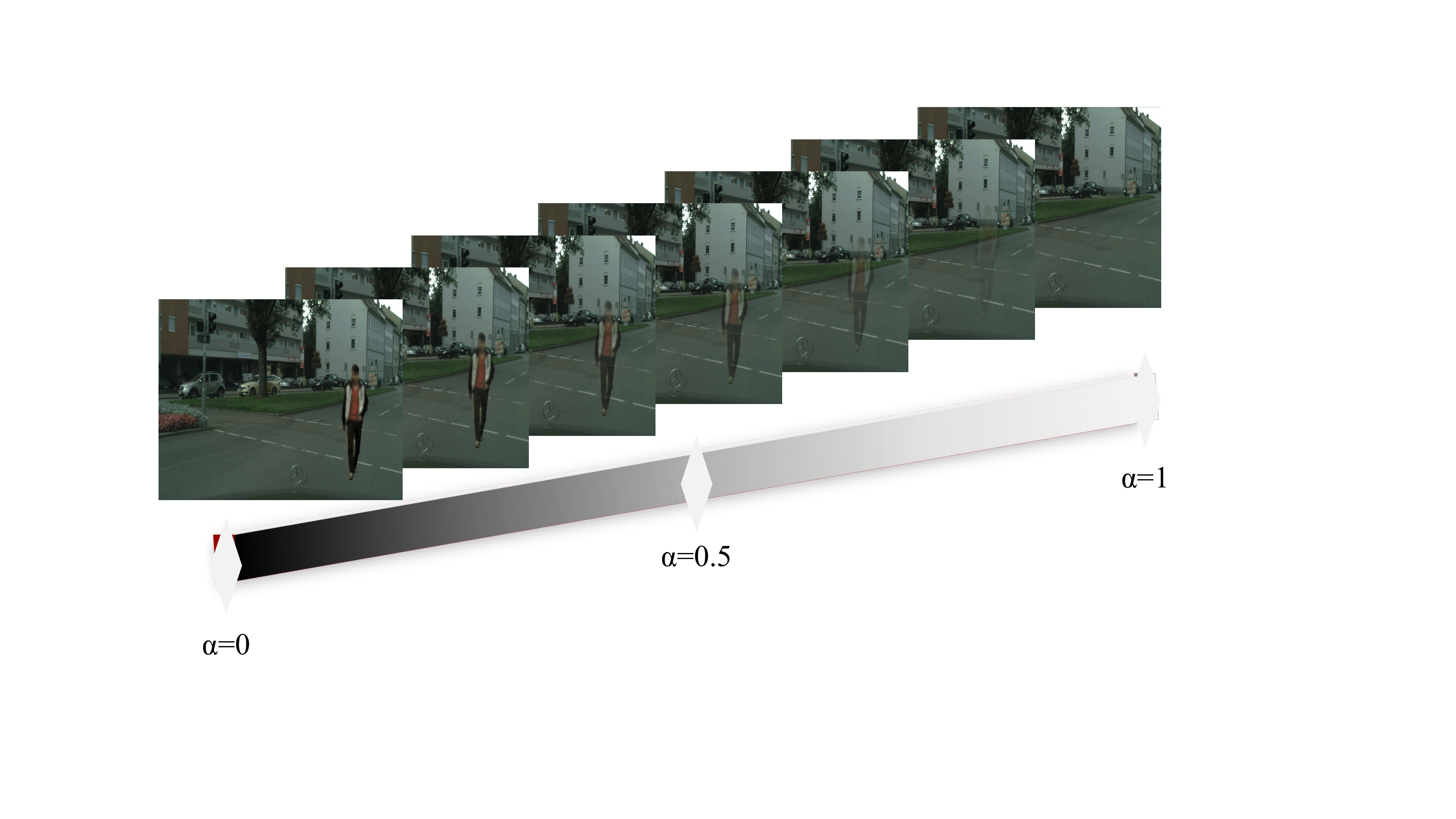}
	\caption{An illustration of $\alpha$ blending for local image degradation effect.}
	\label{fig:alpha_blending}
\end{figure}

We analyse that the reason for the existing learning framework not behaving good lies in the lack of input and output image pair.
In our method, we are able access the original, target and mask images simultaneously with the help of our data synthesis method.
As a result, we are able to propose a new learning framework to boost person removal accuracy as shown in Fig.~\ref{fig:method_compare} (b).

The inspiration mainly comes from some similar tasks, such as watermark removal~\cite{liu2021wdnet,liang2021visible}, raindrop removal~\cite{liang2022single}, and fog removal~\cite{fuh2022mcpa}.
These tasks do not or are hard to obtain the object mask $m$, and the source image $I$ is obtained through $\alpha$ blending, \textit{i.e.},
\begin{equation}
	I = \alpha T + (1-\alpha) O,
\end{equation}
where $O$ is the watermark or raindrop to be removed. 
The removing process is like local image degradation with an inverse $\alpha$ blending process, and the purpose of neural network is to fade the specific object area.
An illustration of $\alpha$ blending is shown in Fig.~\ref{fig:alpha_blending}.
As for person removal, we can consider it as a special case, \textit{i.e.}, $\alpha = 1$.
Thus, image person removal can be regarded as a local image degradation process from state $\alpha=0$ to state $\alpha=1$.

In this way, we are able to perform person removal in a new learning framework.
The source image $I$ is directly fed into the person removal network, and the predictive target image can be represented by the following equation
\begin{equation}
	\hat{T} = (1-m) \otimes I + m \otimes G(I,m),
\end{equation}
where $G$ denotes the person removal network, and we use the watermark removal network proposed in~\cite{liang2021visible} as an example.
By doing this, we are able to retain the color information to the maximum extent in the process of person removal. More closely, the mask can be used to guide the training process to show where the person locates.

\subsection{Coarse-to-fine learning}

Images obtained by the final prediction often have blurring and ghosting area. 
Fortunately, the result can be refined by a training process from coarse to fine.
That is to say, the predictive target image for the first time can be used as the source image in the second time, resulting in the following generation process
\begin{equation}
	\hat{T} = (1-m) \otimes I + m \otimes G(\hat{T},m),
\end{equation}
where $I$ is replaced with $\hat{T}$ in the first time.
By this mean, the effect of $\alpha$ blending is gradually faded, and the target image can be recovered with a good quality.

\section{Experimental analysis}~\label{sec:5}

In this section, we perform experimental analysis to verify the effectiveness of the proposed data synthesis as well as the image person removal framework.

\subsection{Experimental setting}

\textbf{Datasets}. In order to verify the effectiveness of our data synthesis method, we compare the synthesized person removal dataset with existing common object removal dataset, \textit{i.e.}, the Place2~\cite{zhou2017places} and Paris StreetView dataset~\cite{doersch2012makes}.
As for our own data, the background images are selected from Cityscapes~\cite{cordts2016cityscapes} and BDD100K~\cite{yu2020bdd100k} datasets. 
For the image mosaic method, the person models are selected from Daimler Pedestrian Detection Benchmark~\cite{flohr2013pedcut}.
For the virtual rendering method, the person models are selected from PersonX engine~\cite{sun2019dissecting}.
We randomly select 10 person models and 50 background images to generate a dataset with 500 images, in which 70\% is for training and 30\% is for testing.
In the virtual engine, we've designed 15 illumination angles.

\textbf{Models}. We've selected the watermark removal baseline network SLBR~\cite{guo2021image} as our person removal network within our proposed learning framework. SLBR is a CNN-based method and it takes U-Net as the backbone network.
Comparatively, we've selected several state-of-the-art common object removal networks including MEDFE~\cite{liu2020rethinking}, CTSDG~\cite{guo2021image}, BAT~\cite{yu2021diverse} and \textit{et al.} to compare with our method. There methods are commonly GAN-based methods, which apply an adversarial training framework.

\textbf{Evaluation metrics}. Similar to~\cite{liu2020rethinking,yu2021diverse,guo2021image}, we use peak signal-to-noise ratio (PSNR), structural similarity (SSIM), learned perceptual image patch similarity (LPIPS), root mean square error (RMSE) and weighted root mean square error (RMSEw) to evaluate the quality of person removal task.

\subsection{Quantitative analysis}

\begin{table*}[t]
	\caption{Quantitative comparison between different methods. Results are all tested on the real synthesis dataset.}
	\centering
	\begin{tabular}{p{2.2cm}|p{2.4cm}|p{1.2cm}|p{1.0cm}p{1.0cm}p{1.0cm}p{1.0cm}p{1.0cm}}
	\hline
		Method  & Dataset & Images  & PSNR$\uparrow$ & LPIPS$\downarrow$  & SSIM$\uparrow$ & RMSE$\downarrow$ & RMSEw$\downarrow$  \\
	\hline
	\hline
		MEDFE~\cite{liu2020rethinking}   & Paris StreetView & 14.9k & 41.08 & 0.0037 & 0.9969 & 2.69 & 36.41 \\
		BAT~\cite{yu2021diverse}   & Paris StreetView & 14.9k & 42.58 & 0.0018 & 0.9978 & 2.48 & 34.17  \\
		CTSDG~\cite{guo2021image}   & Paris StreetView & 14.9k & 46.51 & 0.0039 & 0.9976 & 1.42 & 19.65 \\
	\hline
		MEDFE~\cite{liu2020rethinking}   & Place2 & 1.8 m & 41.05 & 0.0043 & 0.9967 & 2.76 & 37.20 \\
		BAT~\cite{yu2021diverse}   & Place2 & 1.8 m & 43.48 & \textbf{0.0015} & 0.9980 & 2.23 & 30.54 \\
		AOT-GAN~\cite{zeng2022aggregated}   & Place2 & 1.8 m & 44.54 & 0.0046 & 0.9974 & 1.80 & 30.15 \\
		CTSDG~\cite{guo2021image}   & Place2 & 1.8 m & 47.51 & 0.0029 & 0.9980 & 1.30 & 17.93 \\
		RFR~\cite{li2020recurrent}   & Place2 & 1.8 m & 48.93 & 0.0021 & 0.9985 & 1.14 & 18.76 \\
	\hline
	MEDFE~\cite{liu2020rethinking} & Real synthesis & 0.35k & 40.10 & 0.0040 & 0.9962 & 2.97 & 40.50\\
	AOT-GAN~\cite{zeng2022aggregated} & Real synthesis & 0.35k & 43.90 & 0.0059 & 0.9971 & 1.90 & 28.10\\
	CTSDG~\cite{guo2021image}  & Real synthesis & 0.35k & 47.43 & 0.0028 & 0.9980 & 1.33 & 18.26\\
	\hline
		Ours-R   & Real synthesis & 0.35k & 48.13 & 0.0029 & 0.9981 & 1.24 & 16.54 \\
		Ours-V-Fix   & Virtual synthesis & 0.35k & 48.18 & 0.0027 & 0.9981 & 1.25 & 16.65 \\
		Ours-V-Learn   & Virtual synthesis & 0.35k & 49.13 & 0.0028 & 0.9981 & 1.12 & 14.71 \\
		Ours-V-Full   & Virtual synthesis & 5.25k & \textbf{49.85} & 0.0022 & \textbf{0.9983} & \textbf{1.08} & \textbf{14.23} \\
	\hline
	\end{tabular}
	\label{tab:quantitative_compare}
\end{table*}

\begin{figure*}[t]
	\centering
	\includegraphics[scale=0.6]{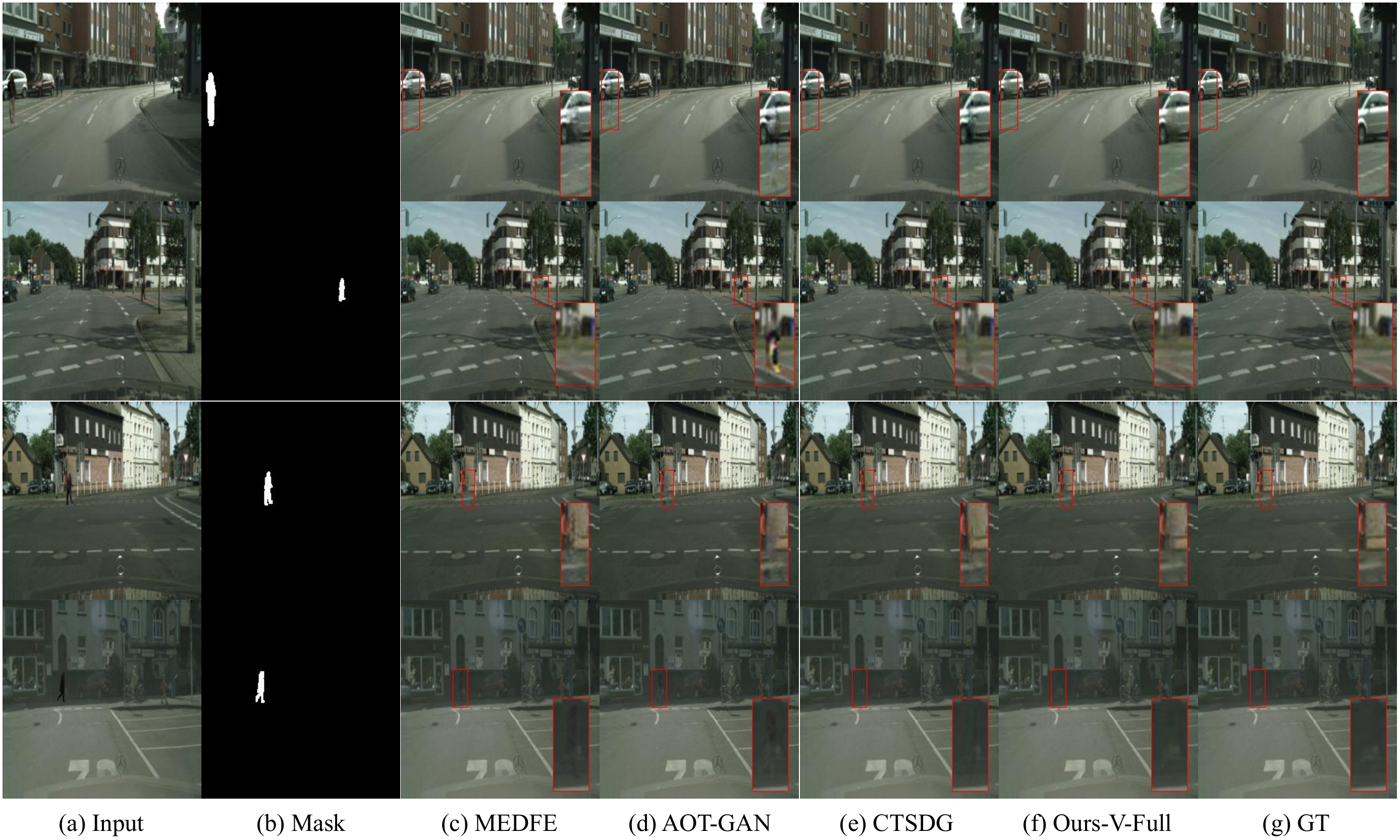}
	\caption{Visualization results of the comparison between existing methods and ours.}
	\label{fig:viz_method_compare}
\end{figure*}

We first perform quantitative comparison with existing methods.
The results are summarized in Table.~\ref{tab:quantitative_compare}, in which existing methods are all trained on large-scale object removal datasets such as Place2~\cite{zhou2017places} with 1.8 million images and Paris StreetView~\cite{doersch2012makes} with 14.9k images or on our dataset, while our method is trained all on the real or virtual synthesis dataset proposed in this paper.
All the evaluation metrics are calculated upon the test part of the real synthesis dataset, since the ground truth masks are available here.

Existing methods~\cite{liu2020rethinking,yu2021diverse,guo2021image} trained on Paris StreetView dataset achieve comparable performance against our method.
Among them, CTSDG~\cite{guo2021image} obtains the highest PSNR result (46.51\%) and BAT~\cite{yu2021diverse} gets the lowest LPIPS result (0.0039).

Relatively speaking, the corresponding methods trained on Place2 dataset obtain better results.
For example, BAT~\cite{yu2021diverse} trained on Place2 achieves 43.48\% PSNR compared with 42.58\% on Paris StreetView. CTSDG~\cite{guo2021image} achieves 47.51\% PSNR on Place2 compared with 46.51\% on Paris StreetView.
The reason may be that Place2 has a large dataset size, and the models trained on it can cover more situations.
Among them, RFR~\cite{li2020recurrent} obtains the highest PSNR result (48.93\%) and BAT~\cite{yu2021diverse} obtains the lowest LPIPS result (0.0015).

We also fine-tune the pretrained network on our real synthesis dataset, but the precision does not meet the expectation. Models trained on our dataset behaves worse than those trained on large-scale datasets.
This is, on the one hand, due to the influence of over-fitting, and more importantly these methods directly subtract the person areas, which does not make full use of the data.

As for our method, we train the network on the synthesis dataset.
Network trained on real synthesis dataset (Ours-R) obtains 48.13\% PSNR and 0.0029 LPIPS, which is quite a competitive results against existing methods.
When training on the virtual synthesis dataset, our method achieves 48.18\% PSNR on the fixed lighting condition (Ours-V-Fix), 49.13\% on the learned lighting condition (Ours-V-Learn), and 49.85\% on the full lighting condition (Ours-V-Full).
Networks trained on virtual synthesis dataset obtains better results compared with real synthesis dataset even though they are all test on real synthesis dataset, which effectively verifies the importance of changing lighting condition in virtual environment.
The models trained on virtual synthesis datasets have good generalization ability to real data.

It's also worth mention that the training data used in our method is much smaller than existing object removal datasets.
The real synthesis dataset contains only 350 training samples, and the virtual synthesis dataset contains 5250 training samples at most.
Correspondingly, Place2 and Paris StreetView contain 1.8 million and 14.9k training samples respectively.
We've made these datasets at a small cost but got comparable performance, which is quite a existing result.
We believe the performance can be improved if more efforts can be put on the data synthesis process.

From the viewpoint of network structure, existing methods~\cite{liu2020rethinking,yu2021diverse,guo2021image,zeng2022aggregated,li2020recurrent} follow a common image inpainting framework, in which they first subtract the person areas and then recover the images.
Such a training process may lead to the loss of important texture information during feature extraction.
As a result, they are hard to fill the missing parts of the image well.
As for our method, we first produce a specific person removal mask with the help of data synthesis, and then a mask-guided network is trained with a coarse-to-fine training process.
During this stage, the person area in the image is slowly faded like a $\alpha$ blending operation.
And thus, we are able to grasp the texture information during the training process.
The network and data synthesis method perfectly fit together to such a task, and that's why we are able to obtain the best results among all.


\subsection{Qualitative analysis}

\begin{figure*}[t]
	\centering
	\includegraphics[scale=0.6]{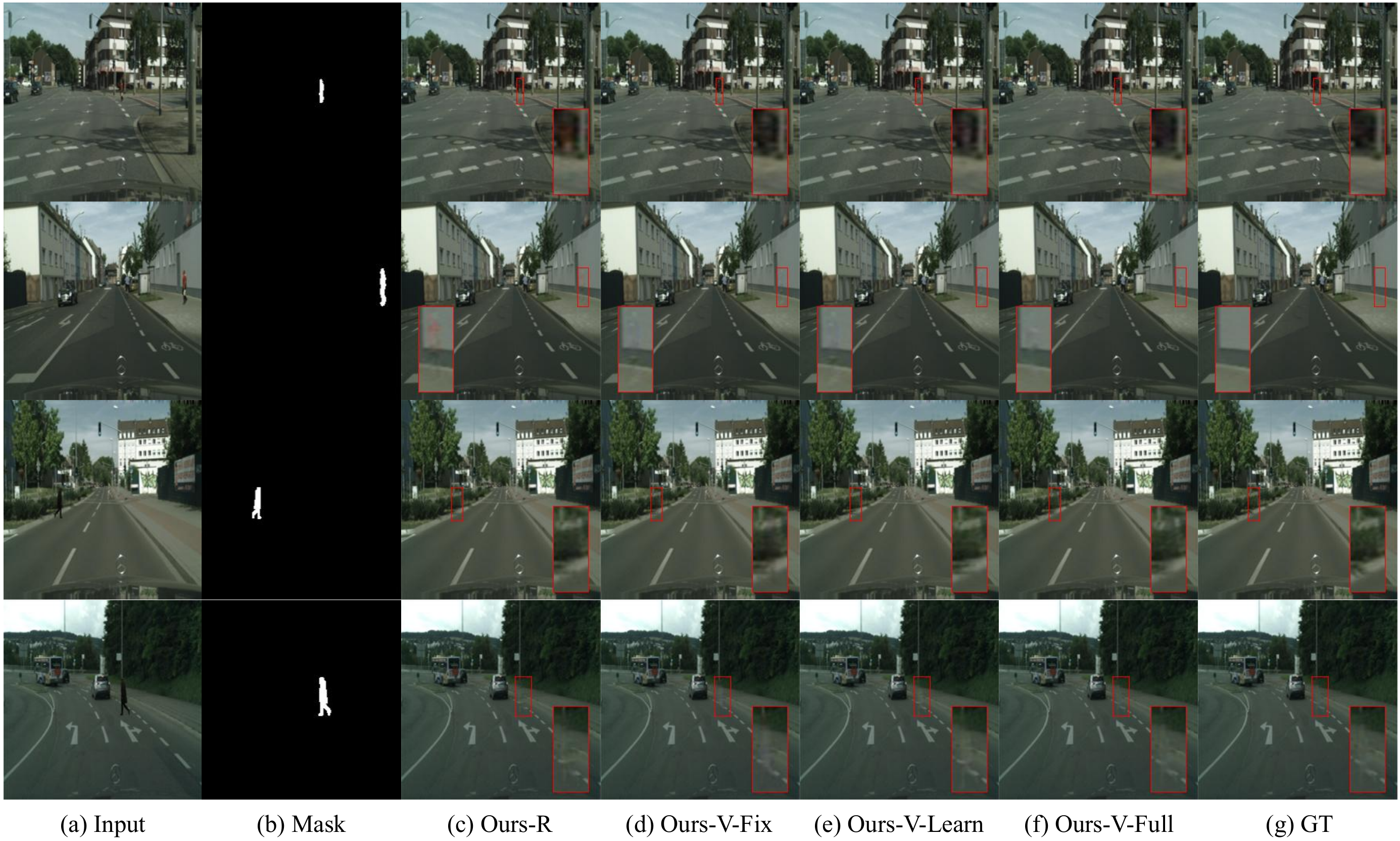}
	\caption{Visualization results of the comparison between the networks trained on different synthesis datasets.}
	\label{fig:data_compare}
\end{figure*}

\begin{figure*}[t]
	\centering
	\includegraphics[scale=0.6]{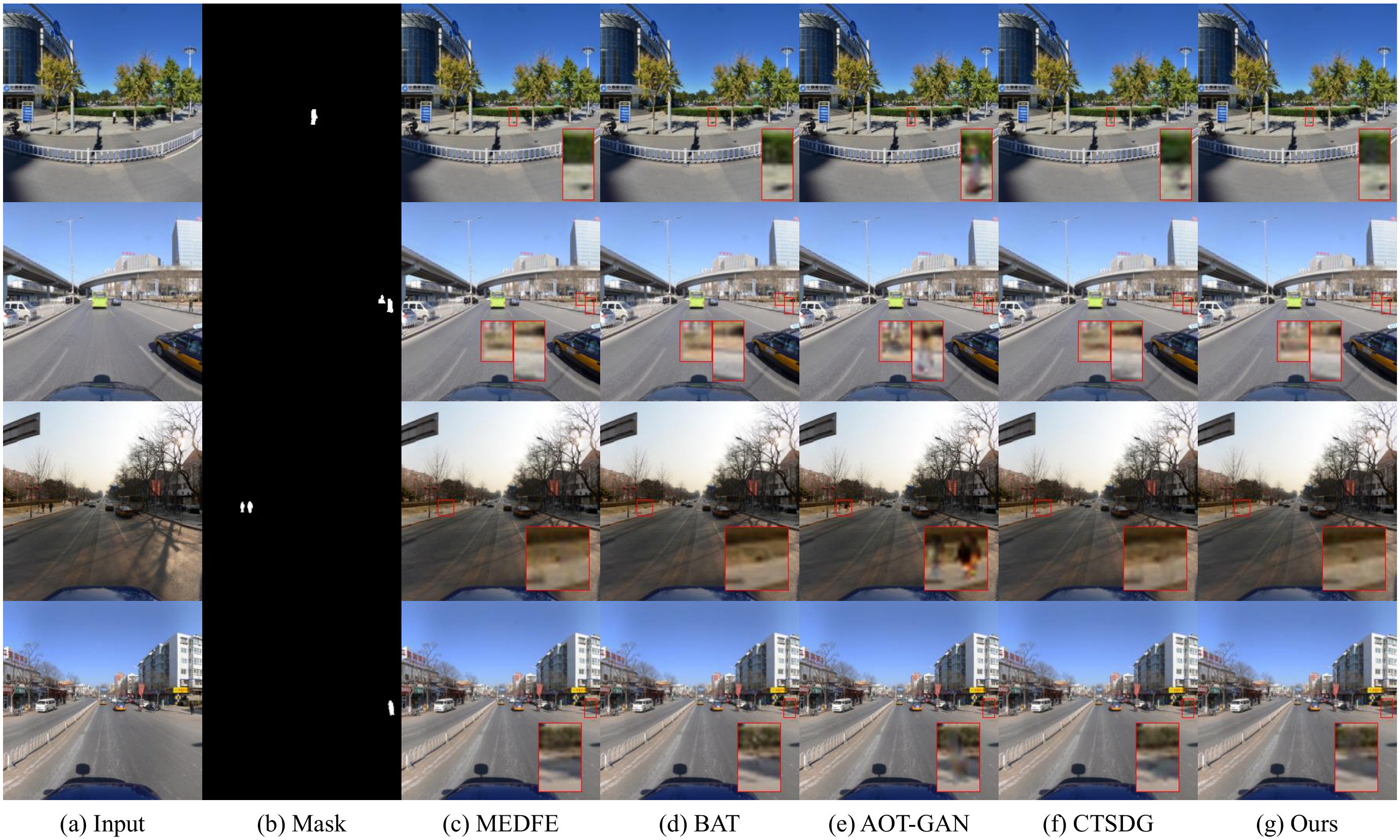}
	\caption{Visualization results of different methods testing on third-party unseen images.}
	\label{fig:other_data}
\end{figure*}

The corresponding qualitative comparison between different methods are presented in Fig.~\ref{fig:viz_method_compare}.
Existing methods including MEDFE~\cite{liu2020rethinking}, AOT-GAN~\cite{zeng2022aggregated} and CTSDG~\cite{guo2021image} are trained on Place2 dataset, and our method is trained on the virtual synthesis dataset with full lighting condition (Ours-V-Full).
Despite training on large-scale dataset, existing methods focus on common object removal, and they behave not so good on the person removal task.
Severe blurring and ghosting are generated since they do not take color information into account during the training process and the texture information is ignored.
For example, AOT-GAN~\cite{zeng2022aggregated} behaves bad on removing person with colourful texture. MEDFE~\cite{liu2020rethinking} and CTSDG~\cite{guo2021image} contain image blurs in the recovery area.
Our method is trained on the specific person removal dataset, and apply the mask to guide the feature extraction process, so that the texture features are kept to some extent.
As a result, out method obtains the best visual effect compared with existing methods.

We also compare the networks trained on different synthesis datasets in our method. The corresponding visualization results are shown in Fig.~\ref{fig:data_compare}.
Although tested on real synthesis dataset, the visual effect of networks trained on virtual synthesis datasets are much better.
Fewer image blurs and artifacts are obtained.
An obvious observation is that network trained on real synthesis dataset (Ours-R) behaves terrible in removing person area with bright colors, which can be seen from the first and second rows.
Another observation is that virtual synthesis dataset with fixed lighting condition (Ours-V-Fix) is hard to remove dark person area, because the dark person appears rarely in this case, which can be seen from the third and fourth rows.
Virtual synthesis datasets with learned and full lighting condition perform well, and the results are close to the ground truths.

The previous experiments are tested on the dataset generated by our data synthesis method.
In order to verify the robustness of our method, we conduct a comparative test of different methods on a set of third party unseen images.
The images are all collected from existing benchmark datasets, and we manually generate some person masks.
The visualization results are shown in Fig.~\ref{fig:other_data}, from which we can see that our method obtains the best visual effect compared with other methods.
Among them, AOT-GAN~\cite{zeng2022aggregated} behaves the worst and there exist strong image blurs and artifacts in the image areas.
As for other methods such as MEDFE~\cite{liu2020rethinking}, BAT~\cite{yu2021diverse} and CTSDG~\cite{guo2021image}, the recovery effect is comparable to ours. However, the shadow areas in the recovery areas are slightly larger than ours.
Our method is better than the existing method in retaining some details.

\subsection{Ablation study}

\begin{table}[t]
	\caption{Ablation study on the proposed network structure. `CM' denotes color maintaining. `MG' denotes mask guidance. `CF' denotes coarse-to-fine training stage.}
	\centering
	\begin{tabular}{p{1.2cm}|p{1.0cm}p{1.0cm}p{1.0cm}p{1.0cm}p{1.0cm}}
	\hline
		Method   & PSNR$\uparrow$ & LPIPS$\downarrow$  & SSIM$\uparrow$ & RMSE$\downarrow$ & RMSEw$\downarrow$  \\
	\hline
	\hline
		w/o CM    & 48.37 & 0.0031 & \textbf{0.9980} & 1.22 & 16.13 \\
		w/o MG    & 48.94 & 0.0026 & 0.9981 & 1.22 & 16.12 \\
		w/o CF    & 49.44 & 0.0023 & 0.9983 & 1.17 & 15.31 \\
	\hline
		Ours     & \textbf{49.85} & \textbf{0.0022} & 0.9983 & \textbf{1.08} & \textbf{14.23} \\
	\hline
	\end{tabular}
	\label{tab:ablation_study}
\end{table}

We perform ablation study on the important component of the proposed network structure.
The results are summarized in Table~\ref{tab:ablation_study}.
`CM' denotes color maintaining, which means the person area is not subtracted and the RGB information is kept. Without CM, the performance degrades. This verifies the importance of the texture information for image person removal during training.
`MG' denotes mask guidance, which means the mask is concatenated with the image before sending into the CNN extractor.
Without MG, the performance also drops, which verifies the effectiveness of the mask-guided network structure.
`CF' denotes coarse-to-fine training stage, which means the intermediate output of the first stage is fed into the second stage to refine the prediction.
Without CF, the performance also degrades.
This significantly verifies the effectiveness of the refining stage.

\subsection{Limitation}

\begin{figure}
	\centering
	\includegraphics[scale=0.65]{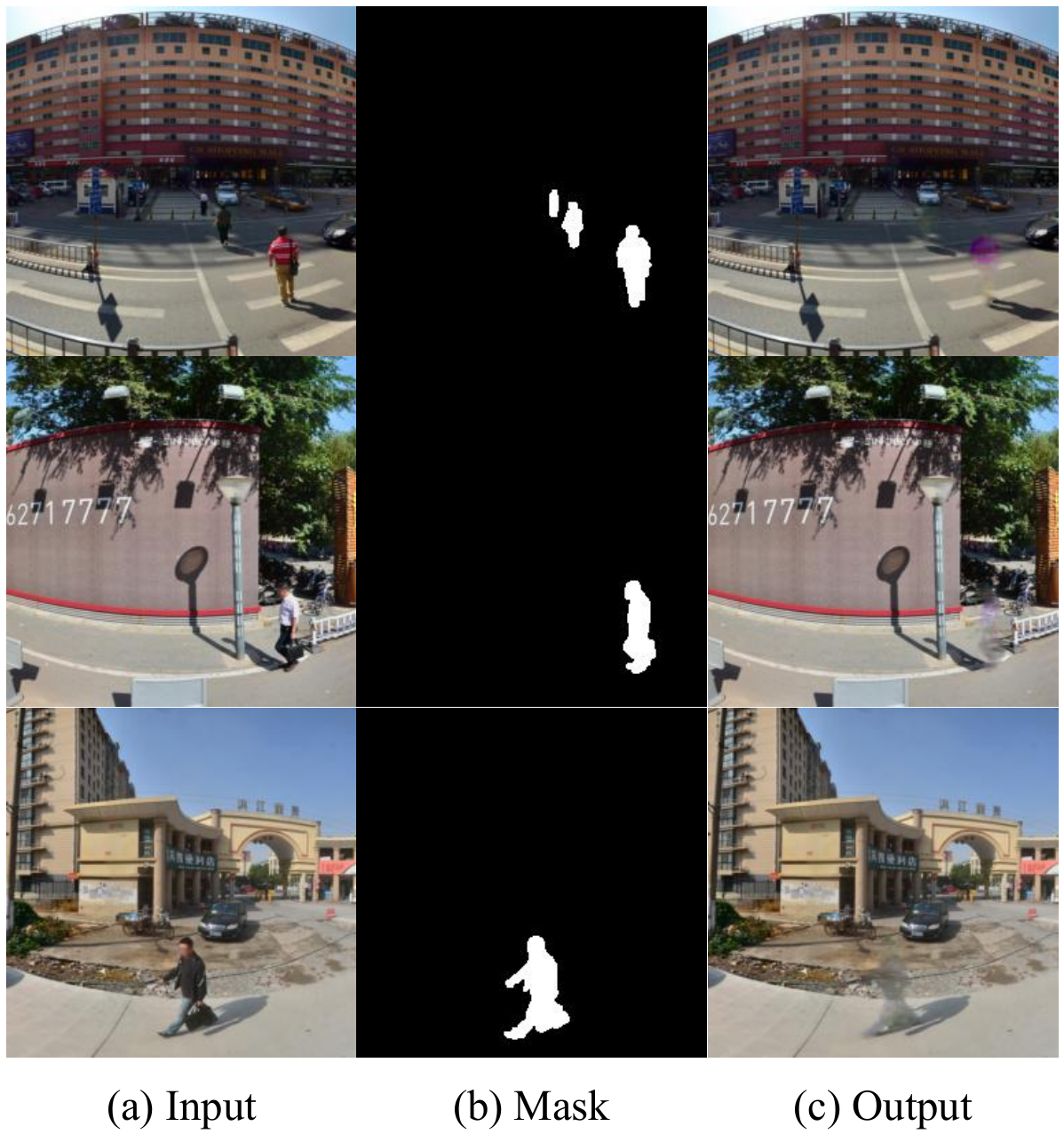}
	\caption{Failure cases of person removal in our method.}
	\label{fig:failure_case}
\end{figure}

Although our method performs well on the current dataset, some problems are also found during the third-party image test.
The results are presented in Fig.~\ref{fig:failure_case}.
The first row shows that our method is hard to handle large person area with bright clothes, resulting in blurring images.
The second and third rows indicate that there exist strong image blurs for the method to removing large person area.
Despite the results are somewhat disappointing, we guess that the dataset we've made from 10 person models and 50 background images has many missing modes, which may cause the monotonicity of the training data.
In the next step, we will expand our dataset to include more situations to reduce the impact of the above problems.

\section{Conclusion}

Image person removal aims to remove specific person area in the image, which is one of the sub-tasks for common object removal.
There is no existing person removal dataset yet.
In this paper, we propose two types of dataset synthesis methods to generate a person removal dataset quickly.
Fine person masks are generated simultaneously.
With the help of synthesis dataset, we design a mask-guided person removal network with a coarse-to-fine training process.
The proposed network fits well with the proposed dataset synthesis method in this person removal task. As a result, we obtain the best results quantitatively and qualitatively.
However, our method is limited to remove small person area, which may be due to the problem of raw materials for dataset synthesis.
In next step, we will move to use more accurate models to generate training samples, and try to improve the removal effect of large person area through modifying the network structure.

\bibliographystyle{IEEEtran}
\bibliography{refs}

\end{document}